\documentclass[conference]{IEEEtran}
\IEEEoverridecommandlockouts
\usepackage{cite}
\usepackage{amsmath,amssymb,amsfonts}
\usepackage{algorithmic}
\usepackage{graphicx}
\usepackage{textcomp}
\usepackage{xcolor}
\usepackage{soul}
\usepackage{multirow}
\usepackage{tabu}                      
\usepackage{booktabs}                  
\usepackage{lipsum}                    
\usepackage{mwe}                       
\usepackage{mathptmx}   
\usepackage{subcaption}
\def\BibTeX{{\rm B\kern-.05em{\sc i\kern-.025em b}\kern-.08em
    T\kern-.1667em\lower.7ex\hbox{E}\kern-.125emX}}

\newcommand{\RNum}[1]{\expandafter{\romannumeral #1\relax}}

\usepackage{enumitem}
\usepackage{soul}
\begin{document}

\title{Predicting 3D Motion from 2D Video for Behavior-Based VR Biometrics}

\author{\IEEEauthorblockN{Mingjun Li$^{1}$, Natasha Kholgade Banerjee$^{2}$, Sean Banerjee$^{2}$}
\IEEEauthorblockA{$^{1}$Clarkson University, Potsdam, NY, USA \\
$^{2}$Wright State University, Dayton, OH, USA \\
\texttt{mingli@clarkson.edu}, 
\texttt{\{natasha.banerjee, sean.banerjee\}@wright.edu}}
}

\maketitle

\begin{abstract}
Critical VR applications in domains such as healthcare, education, and finance that use traditional credentials, such as PIN, password, or multi-factor authentication, stand the chance of being compromised if a malicious person acquires the user credentials or if the user hands over their credentials to an ally. Recently, a number of approaches on user authentication have emerged that use motions of VR head-mounted displays (HMDs) and hand controllers during user interactions in VR to represent the user's behavior as a VR biometric signature. One of the fundamental limitations of behavior-based approaches is that current on-device tracking for HMDs and controllers lacks capability to perform tracking of full-body joint articulation, losing key signature data encapsulated by the user articulation. In this paper, we propose an approach that uses 2D body joints, namely shoulder, elbow, wrist, hip, knee, and ankle, acquired from the right side of the participants using an external 2D camera. Using a Transformer-based deep neural network, our method uses the 2D data of body joints that are not tracked by the VR device to predict past and future 3D tracks of the right controller, providing the benefit of augmenting 3D knowledge in authentication. Our approach provides a minimum equal error rate (EER) of 0.025, and a maximum EER drop of 0.040 over prior work that uses single-unit 3D trajectory as the input. 
\end{abstract}

\begin{IEEEkeywords}
Virtual Reality, Biometrics, Authentication, Video Tracking
\end{IEEEkeywords}

\section{Introduction}
\label{sec:intro}

The ability to create immersive virtual reality (VR) applications in mission critical domains such as healthcare~\cite{desselle2020augmented,munoz2022immersive,mehrabi2022immersive,karaosmanoglu2022canoe}, education~\cite{minnekanti2024classesinvr,gao2021digital,bozkir2021exploiting,hasenbein2022learning}, military~\cite{gawlik2020factors,de2020use,ahir2020application,rettinger2022defuse}, and finance~\cite{campbell2019uses,weise2016virtual}, enables users to perform sensitive activities, such as taking an exam, paying a bill, teleoperating a robot, or checking medical records, raises challenges with user security. Given the potential for misuse of user data, there is a growing concern related to ensuring that VR applications are safe from malicious access~\cite{odeleye2023virtually}. Early work in securing a VR application investigated traditional PIN, password, or multi-factor approaches~\cite{alsulaiman2008three,alsulaiman2006novel,gurary2017leveraging,george2020gazeroomlock,yu2016exploration,george2017seamless,olade2020exploring,funk2019lookunlock,george2019investigating} adopted from smartphone, desktop, or laptop-based systems. However, these techniques pose a security risk as the user's account is immediately compromised if an attacker gains access to the credentials. Further, PIN, password, or multi-factor-based approaches do not provide any protection against deliberate attacks by a genuine user. E.g., a patient being asked to perform an at-home VR-based rehabilitation activity providing their credentials in the system and handing the device to an able ally.

To overcome the limitations of traditional approaches for authentication, a body of work has emerged over the past decade that uses the behavior of the person in the VR space as a signature, i.e., to perform behavior-based VR biometric authentication~\cite{mustafa2018unsure,mmm2019vr,pfeuffer2019behavioural,ajit2019combining,miller2019,mathis2020knowledge,mathis2020rubikauth,mathis2020fast,miller2020within,olade2020biomove,miller2020personal,miller2021using,liebers2021understanding,rack2023alyx,liebers2024identifying,li2024using,liebers2024kinetic}. Reviewed at length in a number of surveys~\cite{giaretta2022security,stephenson2022sok,odeleye2023virtually}, the approaches have investigated conducting authentication by using one or more of the headset motion, hand controller movements, and eye gaze patterns while users conduct activities such as ball-throwing, pointing, pinching, object selection, object movement, and gameplay.  

Thus far, all approaches to perform movement-based authentication in VR use the 3D trajectory collected from the VR devices as a signature of the user's behavior to authenticate the user, yielding a sparse set of tracked data concentrated at the hand, headset, and/or eyes. For several VR interactions, users perform whole body motion resulting in multiple joint articulations, which though serving as a signature potential, are lost since current VR systems do not conduct tracking beyond the headset, hands, and eyes. 

We investigate the potential of using the external video of a user to provide knowledge of the person's body articulation as they conduct activities in the VR space. We use video data to track the user's 2D joint articulation using OpenPose~\cite{cao2017realtime}, a standard body joint tracking method. Our work takes advantage of the movement of multiple joints not tracked by the VR device tracking method. Though one approach is to use the 2D tracks directly, in this work, we use the tracks to predict the 3D trajectory of the right controller using a Transformer network~\cite{vaswani2017attention}, to acquire the benefit of three-dimensional knowledge for authentication using deep neural architectures. We augment the knowledge by predicting past and future right controller 3D motion from an input 2D trajectory segment, to leverage the improvement of authentication using future motion forecasting as shown in Li et al.~\cite{li2024using}. 

Upon evaluating the equal error rate (EER), a standard biometric authentication evaluation metric~\cite{jain2007handbook}, using data from the dataset of Miller et al.~\cite{miller2022combining}, we show that our work outperforms baseline methods that eliminate 2D joints not tracked by the VR system, as well as methods that predict current and future 2D trajectories instead of 3D. We obtain the lowest EER of 0.025. Our work outperforms the state-of-the-art method of Li et al.~\cite{li2024using} that uses traditional 3D device trajectories to conduct improved authentication by forecasting 3D output. We show an average EER drop of 0.025 over Li et al.~\cite{li2024using} and a maximum drop of 0.040. Our work shows the benefit of using 2D data on non-tracked joints from video in improving behavior-based authentication in VR.

\section{Related Work}
A large body of work has emerged over the last decade to enable VR authentication, summarized in multiple recent reviews~\cite{giaretta2022security,stephenson2022sok,odeleye2023virtually}. Among the first approaches to VR authentication has been porting of traditional credentials such as PIN and password, and their adaptation to VR environments, e.g., via object arrangements~\cite{george2019investigating,george2020gazeroomlock}. Though some resistance to shoulder-surfing has been shown, successful malicious access of the credential compromises security. A number of recent approaches investigate the use of user interactions with VR headsets and hand controllers as a behavior signature, with work spanning the use of individual or combination of headset motion, hand movements, and eye gaze. Recent work uses deep neural networks~\cite{mathis2020knowledge,miller2021using,miller2022combining,liebers2021using,liebers2021understanding,li2024evaluating,li2024using,liebers2024kinetic} based on their ability to represent non-linearities in the data. Various activities have been investigated, including ball-throwing~\cite{miller2022combining}, pointing~\cite{pfeuffer2019behavioural}, door-opening~\cite{li2024evaluating}, pinching~\cite{suzuki2023pinchkey}, object movement~\cite{mathis2020knowledge}, and gameplay~\cite{liebers2023exploring}. 

Prior methods use movements as tracked by the tracking mechanisms available on existing VR devices. In devices such as the HTC VIVE that use lighthouses, IR light emitted by the lighthouse is tracked throughout performance by receptors on the headset and hand controllers, resulting in no more than 3 tracks, or 4 tracks in headsets that track eye gaze. Camera-based tracking such as on the Meta Quest has traditionally tracked the headset using an inside-out approach and the hand controllers through visual observations, resulting in a similar set of 3 or 4 tracks. Datasets used by nearly all approaches thus far have involved participants using hand controllers. Suzuki et al.~\cite{suzuki2023pinchkey} use controller-free hand tracking in the Meta Quest to obtain higher-resolution motion tracks for the fingers, enabling the use of pinching as a signature. However, while datasets like Alyx~\cite{rack2023alyx} exist, none of the approaches has external video data or use full-body joint articulation, e.g., at the elbows and shoulders, the torso, or the lower body, due to their reliance solely on the data tracked by the VR devices that are currently incapable of tracking full-body articulation, only Miller et al.~\cite{miller2022combining} provides external video recordings.

Our method provides the first approach for authentication using external videos to acquire detailed knowledge of full body articulation, while taking advantage of 3D information through intermediate prediction of the 3D controller motion. In conducting knowledge-augmentation by incorporating future 3D controller trajectory prediction using deep neural networks, our method is closest to the work of Li et al.~\cite{li2024using}. However, our work differs in that Li et al.~\cite{li2024using} use VR device 3D trajectory data as input similar to existing methods.

\section{Data Preparation}
\label{sec:data}

\paragraph{Dataset} We utilize the dataset provided by Miller et al.~\cite{miller2022combining}, which contains comprehensive tracking data of virtual reality (VR) interactions, specifically focused on ball-throwing tasks. The dataset encompasses tracked trajectories for both the VR headset and hand controllers, as well as synchronized external video recordings for 46 participants. These participants performed ball-throwing tasks using three different VR systems: the Meta Quest, HTC VIVE, and HTC VIVE Cosmos. 41 out of the 46 participants identified as right-handed and performed the task primarily with their right hand. Each participant completed two data collection sessions for every VR system, with a minimum gap of 24 hours between sessions to mitigate the potential effects of fatigue or learning bias. In each session, participants performed 10 trials, where each trial consisted of a single ball-throw action. Data was recorded for a duration of three seconds per trial, capturing the full motion sequence of the throw. Additionally, the dataset includes external video footage captured from a side-view angle using a GoPro camera, which complements the motion data by providing an alternative visual perspective of the participants' physical movements during the trials. In this work, we focus on the right-hand controller data from the 41 right-handed participants interacting with the HTC VIVE system. This choice is driven by the fact that, for these participants, the right-hand controller provides the most dominant and representative motion data during the ball-throwing task. The HTC VIVE tracking data was recorded at a frame rate of 45 frames per second (FPS), resulting in a total of 135 frames captured over the 3-second period for each trial. In parallel, the GoPro camera recorded video at 60 FPS, yielding 180 frames over the same duration.


\paragraph{Generating 2D Trajectory from Video}
\begin{figure*}[ht!]
    \centering
    \includegraphics[width=.9\linewidth]{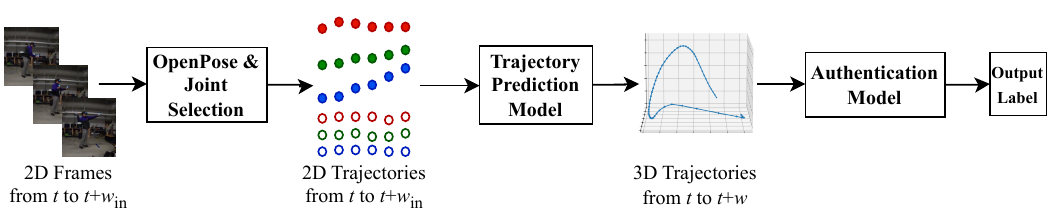}
    \caption{We extract 6 body joints, i.e., right shoulder (solid red), elbow (solid green), wrist (solid blue), hip (hollow red), knee (hollow green), and ankle (hollow blue), by OpenPose from 2D images in time range $t$ to $t+w_\textrm{in}$ (with length $w_\textrm{in}$). We then feed the body joints image coordinates into the forecasting model, which outputs the 3D trajectory sequence from time $t$ to $t+w$ (with length $w$). The 3D predicted trajectory serves as the direct input for the authentication model.}
    \label{fig:model}
\end{figure*}
 
To extract the participants' body movements from the external video recordings, we use OpenPose~\cite{cao2017realtime} to automatically track and obtain 2D skeletal representations for each participant. 
In this work, we specifically select the 2D trajectories corresponding to the right wrist, elbow, shoulder, hip, knee, and ankle. These joints are chosen because the right wrist is anatomically proximal to the right-hand controller used in the VR system, and the other joints are expected to significantly contribute to the mechanics of the ball-throwing action. 
The lower body joints (hip, knee, and ankle) are crucial in stabilizing the participant's posture and contributing to the transmission of force from the lower body to the upper body during the throw. These joints are key in facilitating the rotational movements and weight transfer needed to generate power and control during the throwing motion. The integration of both upper and lower body joint data allows for a more comprehensive analysis of the participant's biomechanics during the ball-throwing task. To ensure consistency in our analysis, we address the discrepancy in the frame rates between the external GoPro camera (60 FPS, yielding 180 frames over a 3-second trial) and the VR device (45 FPS, producing 135 time samples over the same 3-second period). We apply a uniform downsampling process to the 2D skeletal trajectories obtained from OpenPose, aligning them with the temporal resolution of the VR device data. After downsampling, both the 2D joint trajectories and the 3D trajectories from the right-hand controller are synchronized to contain 135 time samples over the 3-second trial period, allowing for direct comparison and analysis.

\paragraph{Sliding Window Data Generation}

We conduct authentication in sliding windows of size $w_\textrm{in}$ over the length of the trajectory. The use of sliding windows enables us to evaluate authentication over variable timespans, shown by the window choices discussed in Sections~\ref{sec:exp} and \ref{sec:results}. For each sliding window of size $w_\textrm{in}$, we extract 2D body joint trajectory segments to create a tensor of dimensions $w_\textrm{in} \times n_\textrm{in} \times 2$ per segment, where $n_\textrm{in}$ represents the number of joints used. In our case, $n_\textrm{in}$ is 6 covering the wrist, elbow, shoulder, hip, knee, and ankle. To take advantage of the findings in Li et al.~\cite{li2024using} that incorporating future trajectory information shows higher authentication success, at the output, we extract right controller 3D trajectory segments over sliding windows of size $w > w_\textrm{in}$, yielding a matrix of size $w \times 3$ per segment. Each 3D trajectory segment of size $w$ starts at the same time stamp as its corresponding 2D segment of size $w_\textrm{in}$. Per choice of $w_\textrm{in}$ and $w$, we extract all possible sliding windows for the genuine user while ensuring that 2D and 3D trajectory segments are obtained from the 135 time samples. We pair each genuine user's trajectory data with trajectory data from multiple randomly selected users, excluding the genuine user, where the randomly selected users serve as impostors. The resulting data enables us to train one trajectory prediction and authentication model per user. For each genuine user trajectory segment, we select an impostor and one session from that impostor at random. We extract a trajectory segment at the same timestamp as that of the genuine user trajectory segment. We repeat the process till the set of impostor trajectory segments is of the same size as the genuine user trajectory set, yielding balanced positive (genuine) and negative (impostor) data per genuine user.

\section{Neural Architecture for Authentication using 3D Trajectory Prediction}
\label{sec:dnn}

The goal of our neural architecture is to take in a set of body joint 2D trajectories over a portion of each window, $w_\textrm{in} \times n_\textrm{in} \times 2$, predict the 3D trajectory of the right controller over the full window, $w \times 3$, and use the predicted 3D trajectory to conduct authentication. Here, $w_\textrm{in} < w$, i.e., from a smaller window of 2D trajectory data, we generate a larger window of 3D trajectory data, including predicting future motion in the region $w-w_\textrm{in}$. As shown by Fig.~\ref{fig:model}, our architecture consists of a trajectory prediction network model, $M_\textrm{traj}$, and an authentication network, $M_\textrm{auth}$.

\paragraph{Trajectory Prediction Model}
\begin{figure}[h!]
    \centering
    \includegraphics[width=.8\linewidth]{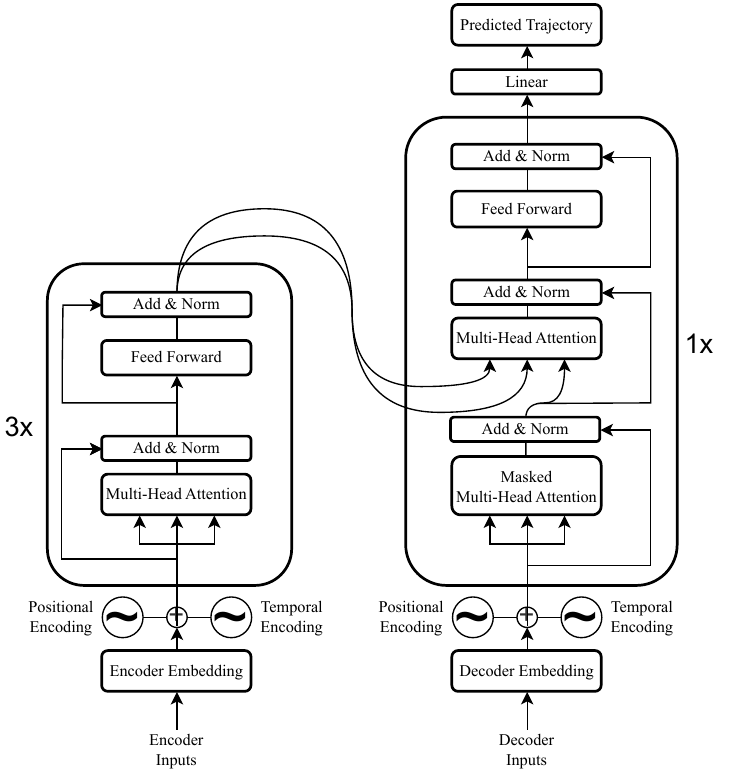}
    \caption{Trajectory Prediction Model.}
    \label{fig:fore_model}
\end{figure}

The trajectory prediction model $M_\textrm{traj}$ takes a 2D body joint trajectory sequence with length $w_\textrm{in}$ as input and predicts the 3D trajectory of length $w$ as output. As shown in Fig.~\ref{fig:fore_model}, $M_\textrm{traj}$ is based on Transformer~\cite{vaswani2017attention}, and is particularly a variant of the Informer~\cite{zhou2021informer} that predicts future time series data from past data. Consistent with encoding methods employed in Transformer and Informer, we perform the following encoding operations to enhance the representative capacity of the raw data:
\begin{enumerate}[noitemsep,topsep=0pt,leftmargin=*]
    \item We employ learned embeddings to elevate the dimensionality of the input 2-dimensional trajectory coordinates across all timestamps, thereby capturing nuanced information not discernible in the original space.
    \item We encode the position of each timestamp within the window, enabling extraction of positional correlations.
    \item We conduct temporal encoding for each timestamp within the global time series, to encompass the entirety of a motion session. As in Informer, the information captures the overarching temporal dynamics in the window.
\end{enumerate}

We aggregate the outputs from the three encoding operations through summation to form the input for the encoder and decoder. The encoder of $M_\textrm{traj}$ uses the same foundational architecture as the Transformer encoder. We use 3 identical encoder layers, with each layer consisting of two sub-layers\textemdash{}a multi-head self-attention mechanism and a position-wise fully connected feed-forward network. We incorporate residual connections and layer normalization at the outputs of each sub-layer. The decoder of the Informer architecture is designed for the purpose of forecasting by taking the latter portion of the input time series and using it to generate a new time series consisting of the latter portion as overlap concatenated with future portions of the time series. The architecture is appropriate when the input and output modalities are the same, e.g., as in the work of Li et al.~\cite{li2024using}, where the future 3D trajectory of the right controller is forecasted from earlier right controller 3D trajectory data. In our scenario, the input modality consisting of multiple 2D joint trajectories differs from the output modality of the 3D right controller trajectory. We modify the Informer decoder in two ways to enable cross-modality output generation. First, instead of using a part of the input trajectory, we use the complete 2D trajectory as the input to the decoder. Second, we predict the entire 3D trajectory in the window $w$ that includes data spanning the entire input window $w_\textrm{in}$, unlike in Informer where the latter portion of the 3D trajectory is predicted with a partial overlap with the input window. We use 1  decoder layer, with multi-head attention and feed-forward sub-layers similar to the encoder.

\paragraph{Authentication Model}
\begin{figure}[h!]
    \centering
    \includegraphics[width=\linewidth]{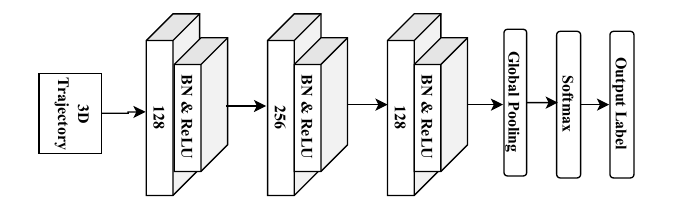}
    \caption{Authentication Model.}
    \label{fig:auth_model}
\end{figure}

The authentication model $M_\textrm{auth}$, as shown in Fig.~\ref{fig:auth_model}, is a fully convolutional network (FCN) for time series data~\cite{wang2017time} that has been shown to be successful in traditional VR biometric authentication using 3D VR headset and controller motions~\cite{mathis2020knowledge,miller2021using}. 
The FCN takes the 3D trajectory predicted using $M_\textrm{traj}$ and predicts a binary label, with 1 indicating that the user is authenticated and 0 otherwise. $M_\textrm{auth}$ consists of three convolution blocks, consisting of a convolutional layer with 1D kernel, batch normalization, and rectified linear unit activation. The convolutional layers provide 128, 256, and 128 outputs each. 
After the 3 blocks, we apply global average pooling 
and softmax to generate the final binary output.

\paragraph{Loss Function}
We estimate weights of $M_\textrm{traj}$ and $M_\textrm{auth}$ by minimizing loss 
\begin{equation}
    \mathcal{L} =  \mathcal{L}_\textrm{traj} + \lambda \mathcal{L}_\textrm{auth}.
    \label{eq:1}
\end{equation}
In Equation~\eqref{eq:1}, $\mathcal{L}_\textrm{traj}$ represents the reconstruction loss of each 3D trajectory of window size $w$ and is given as 
\begin{equation}
 \mathcal{L}_\textrm{traj} = \frac{1}{|\mathcal{W}|} \textstyle\sum_{i \in \mathcal{W}} \textrm{MSE}(\mathbf{T}_i^\textrm{pred}, \mathbf{T}_i^\textrm{gt}),
\end{equation}
where $\mathcal{W}$ represents the total windows, MSE represents the mean-squared error function, and $\mathbf{T}_i^\textrm{pred}$ and $\mathbf{T}_i^\textrm{gt}$ respectively represent the predicted and ground truth trajectories. $\mathcal{L}_\textrm{auth}$ represents the classification loss for the authentication network, and is given as 
\begin{equation}
    \mathcal{L}_\textrm{auth} = \frac{1}{|\mathcal{W}|} \textstyle\sum_{i \in \mathcal{W}} \textrm{BCE}(y_i^\textrm{pred}, y_i^\textrm{gt}),
\end{equation} 
where BCE represents the binary cross-entropy loss function, and $y_i^\textrm{pred}$ and $y_i^\textrm{gt}$ respectively represent the predicted and ground truth class labels, 1 for a genuine user and 0 for an impostor. The factor $\lambda$ weights the loss contributions.

\paragraph{Implementation Details} 
We configure $M_\textrm{traj}$ to have $3$ identical encoder layers and an individual decoder layer. As in the original Transformer architecture, we set the model dimension to 512, with each layer incorporating an 8-head self-attention mechanism. We set the query, key, and value dimensions to 64. We set the dimension of the fully connected layer that generates the output trajectory to 2048. We use filter sizes of $\{128, 256, 128\}$ and 1D kernel sizes of $\{8, 5, 3\}$ for the 3 convolutional blocks of $M_\textrm{auth}$. We set $\lambda$ to 0.5. We optimize using Adam with a learning rate of $1\textrm{e}-4$.

\section{Experiments}
\label{sec:exp}

We conduct authentication experiments in window sizes $w$ ranging from 40 to 130 in steps of 10, and $w_\textrm{in}$ ranging from 30 to 70 in steps of 10. Table~\ref{tab:eer} shows particular choices of $w$ and $w_\textrm{in}$. We perform training using data from the first session of each participant, and perform testing using data from the second session. The two sessions are acquired on two separate days per participant in the Miller dataset. 

Our main authentication results use right controller 3D trajectory predictions from input 2D data for 6 OpenPose joints, i.e., the right wrist, elbow, shoulder, hip, knee, and ankle, resulting in $n_\textrm{in}=6$. We refer to these results as \texttt{3Dfrom2D\_WESHKA}. As a baseline, we compare our approach to Li et al.~\cite{li2024using} who use input 3D trajectory segments to predict an output 3D trajectory containing future forecasted data and prior data with an overlap with the input trajectory. We use an overlap of $w_\textrm{in}/2$ to generate an output trajectory of length $w - w_\textrm{in}/2$. We obtain the full 3D trajectory within the window $w$ for authentication by concatenating the first half of the input trajectory of size $w_\textrm{in}/2$ with the output trajectory. 

We include 10 further experiments in this study, (\RNum{1}) \texttt{2Dfrom2D\_W} evaluates the performance of trajectory generation when predicting just the 2D wrist trajectory within $w_\textrm{in}$, resulting in $n_\textrm{in}=1$. And (\RNum{2}) \texttt{2Dfrom2D\_WES} uses input 2D data from the wrist, elbow, and shoulder in $w_\textrm{in}$ resulting in $n_\textrm{in}=3$. These two approaches mimic video-only biometric authentication. The other two experiments, (\RNum{3}) \texttt{3Dfrom2D\_W} and (\RNum{4}) \texttt{3Dfrom2D\_WES} focus on evaluating the 3D trajectory prediction using 2D data from the upper body. Specifically, \texttt{3Dfrom2D\_W} assesses the predictive capability using 2D data from the right wrist. We have \texttt{3Dfrom2D\_WES} to utilize input data from three joints, the right shoulder, elbow, and wrist. We employ \texttt{3Dfrom2D\_W} to investigate the potential advantages of incorporating multiple joints, as exemplified by \texttt{3Dfrom2D\_WES}, as opposed to relying solely on data from a single track, particularly the track closest to the controller. The subsequent six experiments, denoted as (\RNum{5}) \texttt{3Dfrom2D\_WESH}, (\RNum{6}) \texttt{3Dfrom2D\_WESK}, (\RNum{7}) \texttt{3Dfrom2D\_WESA}, (\RNum{8}) \texttt{3Dfrom2D\_WESHK}, (\RNum{9}) \texttt{3Dfrom2D\_WESHA}, and (\RNum{10}) \texttt{3Dfrom2D\_WESKA}, expand upon the foundation laid by the former experiments by integrating additional body joints from the lower body. Specifically, the letter codes appended to each experiment denote the inclusion of joints at various lower body positions, where `\texttt{H}' represents the hip, `\texttt{K}' denotes the knee, and `\texttt{A}' the ankle. 



\setlength{\tabcolsep}{3.5pt}
\begin{table*}[h!]
  \caption{Average MSE of 3D forecasted trajectories among all participants between the forecasted trajectories and the ground truth trajectories for experiments using 2D as inputs (lower is better).}
  \centering
  \label{tab:fore_mse}
  {%
\begin{tabular}{|c|c|c|c|c|c|c|c|c|c|c|c|c|c|c|c|c|c|c|c|c|}
\hline
$w$ & 40 & 50 & 50 & 60 & 60 & 70 & 70 & 70 & 80 & 80 & 80 & 90 & 90 & 90 & 100 & 100 & 110 & 110 & 120 & 130 
\\ \hline  
$w_\textrm{in}$ & 30 & 30 & 40 & 40 & 50 & 40 & 50 & 60 & 50 & 60 & 70 & 50 & 60 & 70 & 60 & 70 & 60 & 70 & 70 & 70 
\\ \hline

\texttt{3Dfrom2D\_W}  & 0.38 & 0.45 & 0.38 & 0.41 & 0.38 & 0.41 & 0.41 & 0.38 & 0.40 & 0.38 & 0.37 & 0.42 & 0.39 & 0.38 & 0.40 & 0.39 & 0.39 & 0.37 & 0.37 & 0.41
\\ \hline

\texttt{3Dfrom2D\_WES}  & 0.36 & 0.40 & 0.38 & 0.39 & 0.36 & 0.39 & 0.40 & 0.34 & 0.39 & \textbf{0.35} & 0.34 & 0.40 & 0.40 & 0.37 & 0.38 & 0.37 & 0.38 & 0.37 & 0.37 & 0.43
\\ \hline

\texttt{3Dfrom2D\_WESH}  & 0.35 & 0.38 & 0.34 & \textbf{0.36} & 0.36 & 0.38 & 0.37 & 0.34 & 0.38 & 0.36 & 0.35 & 0.39 & 0.38 & \textbf{0.35} & 0.38 & 0.37 & 0.40 & 0.38 & 0.37 & 0.38
\\ \hline

\texttt{3Dfrom2D\_WESK}  & 0.33 & 0.38 & 0.35 & 0.38 & 0.35 & 0.40 & 0.36 & 0.34 & 0.38 & 0.36 & 0.33 & 0.39 & 0.37 & \textbf{0.35} & 0.38 & 0.38 & 0.38 & 0.38 & 0.37 & 0.36
\\ \hline

\texttt{3Dfrom2D\_WESA}  & 0.33 & 0.38 & 0.35 & 0.38 & 0.35 & 0.40 & 0.36 & 0.34 & 0.38 & 0.36 & 0.33 & 0.39 & 0.37 & \textbf{0.35} & 0.38 & 0.38 & 0.38 & 0.38 & 0.37 & 0.36
\\ \hline

\texttt{3Dfrom2D\_WESHK}  & 0.35 & 0.39 & 0.34 & 0.38 & 0.35 & 0.39 & 0.38 & 0.34 & 0.38 & 0.36 & 0.34 & 0.40 & 0.38 & 0.36 & 0.38 & \textbf{0.36} & 0.40 & 0.39 & 0.38 & 0.38
\\ \hline

\texttt{3Dfrom2D\_WESHA}  & 0.34 & 0.38 & 0.35 & \textbf{0.36} & 0.35 & 0.40 & 0.38 & \textbf{0.33} & 0.39 & 0.37 & 0.34 & 0.39 & 0.37 & 0.36 & 0.38 & 0.37 & 0.38 & 0.37 & 0.37 & 0.37
\\ \hline

\texttt{3Dfrom2D\_WESKA}  & 0.33 & 0.37 & 0.34 & 0.37 & \textbf{0.34} & 0.39 & 0.38 & \textbf{0.33} & 0.39 & 0.37 & 0.34 & 0.40 & 0.38 & 0.37 & 0.38 & 0.37 & 0.41 & 0.38 & 0.36 & 0.37
\\ \hline

\texttt{3Dfrom2D\_WESHKA}  & \textbf{0.32} & \textbf{0.35} & \textbf{0.33} & \textbf{0.36} & \textbf{0.34} & \textbf{0.36} & \textbf{0.35} & 0.36 & \textbf{0.37} & 0.37 & \textbf{0.32} & \textbf{0.38} & \textbf{0.35} & \textbf{0.35} & \textbf{0.36} & \textbf{0.36} & \textbf{0.37} & \textbf{0.34} & \textbf{0.35} & \textbf{0.35}
\\ \hline
\end{tabular}}%
\end{table*}

\textbf{Evaluation Metric.} We use the Equal Error Rate (EER), a standard metric for benchmarking authentication systems~\cite{jain2007handbook}, to evaluate our authentication model. The EER corresponds to the point where the false acceptance rate (FAR) equals the false rejection rate (FRR), representing the threshold at which the model is equally likely to incorrectly accept an unauthorized input as it is to wrongly reject a legitimate one. A lower EER indicates superior model performance, reflecting a balanced trade-off between false acceptances and false rejections. In this paper, we report EER values ranging from 0 to 1, rather than expressing them as percentages.

\section{Results}
\label{sec:results}

\subsection{Trajectory Prediction}
In order to facilitate a direct comparison of trajectory prediction accuracy, we present the quantitative results in Table~\ref{tab:fore_mse}, showcasing the mean squared error (MSE) values in each cell for all methods that predict future 3D trajectories based on 2D body joints across different combinations of $w$ and $w_\textrm{in}$. Each row in Table~\ref{tab:fore_mse} corresponds to a specific prediction method, while each column represents a distinct combination of $w$ and $w_\textrm{in}$. From Table~\ref{tab:fore_mse}, we observe that the proposed main method, i.e., \texttt{3Dfrom2D\_WESHKA}, outperforms alternative approaches in the majority of combinations. Notably, in 18 out of the 20 cases, \texttt{3Dfrom2D\_WESHKA} achieves the lowest MSE for the predicted trajectories. This highlights the enhanced accuracy of our main method when leveraging six 2D body joints to predict future 3D trajectories, outperforming other approaches that rely on fewer joints. 

Additionally, it is worth noting that when only a single body joint is used for forecasting, as in the \texttt{3Dfrom2D\_W} method in the first row in Table~\ref{tab:fore_mse}, the average MSE across nearly all combinations of $w$ and $w_\textrm{in}$ is higher compared to other methods, which indicates a much larger forecasting error. A similar pattern is observed when three body joints are used for prediction, as in the \texttt{3Dfrom2D\_WES} method, the second row in Table~\ref{tab:fore_mse}, where the majority of $w$ and $w_\textrm{in}$ combinations also result in higher MSE values than methods utilizing a greater number of body joints. This finding underscores the importance of incorporating additional body joints in trajectory prediction models. The results  demonstrate that using a minimal number of joints tends to result in less accurate predictions, with higher error rates. In contrast, methods that include a larger set of body joints improve the accuracy of future trajectory predictions. By providing the model with more comprehensive motion data, especially from joints that are critical for certain physical activities, throwing a ball in our case, we capture the complex dynamics of human movement. This leads to more reliable and precise estimations of future trajectories, and increases prediction accuracy through the increased number of tracked body joints.

\subsection{Authentication}

\setlength{\tabcolsep}{1.1pt}
\begin{table*}[t!]
  \caption{EER for our approach (bottom row) compared against the baseline, Li et al.~\cite{li2024using}, and other experiments. The value highlighted in bold represents the minimum Equal Error Rate (EER), and the values underscored denote the second lowest EER values among all experiments within the current combination of $w$ and $w_\textrm{in}$. }
  \centering
  \label{tab:eer}
  {%
\begin{tabular}{|c|c|c|c|c|c|c|c|c|c|c|c|c|c|}
\hline
$w$ & $w_\textrm{in}$ & \cite{li2024using} & \texttt{2Dfrom2D} & \texttt{2Dfrom2D} & \texttt{3Dfrom2D} & \texttt{3Dfrom2D} & \texttt{3Dfrom2D} & \texttt{3Dfrom2D} & \texttt{3Dfrom2D} & \texttt{3Dfrom2D} & \texttt{3Dfrom2D} & \texttt{3Dfrom2D} & \texttt{3Dfrom2D}\\
& & & \texttt{\_W} & \texttt{\_WES} & \texttt{\_W} & \texttt{\_WES} & \texttt{\_WESH} & \texttt{\_WESK} & \texttt{\_WESA} & \texttt{\_WESHK} & \texttt{\_WESHA} & \texttt{\_WESKA} & \texttt{\_WESHKA} \\
\hline 
40 & 30 & 0.091 & 0.186 & 0.102 & 0.169 & 0.085 & 0.074 & \underline{0.066} & 0.073 & 0.069 & \textbf{0.060} & 0.070 & \textbf{0.060}\\
\hline 
50 & 30 & 0.085 & 0.184 & 0.107 & 0.159 & 0.082 & 0.063 & 0.056 & 0.054 & 0.049 & \textbf{0.042} & 0.053 & \underline{0.047} \\ \cline{1-1} \cline{3-3} \hline 
50 & 40 & 0.078 & 0.182 & 0.093 & 0.144 & 0.068 & 0.067 & 0.064 & 0.062 & 0.070 & \underline{0.059} & \underline{0.059} & \textbf{0.055} \\  \hline 
60 & 40 & 0.075 & 0.172 & 0.095 & 0.136 & 0.059 & 0.053 & 0.050 & 0.048 & 0.054 & 0.046 & \textbf{0.042} & \underline{0.044} \\ \cline{1-1} \cline{3-3} \hline 
60 & 50 & 0.073 & 0.168 & 0.103 & 0.120 & 0.055 & 0.061 & 0.053 & 0.054 & 0.060 & \underline{0.052} & 0.065 & \textbf{0.051} \\ \cline{1-1} \cline{3-3} \hline 
70 & 40 & 0.070 & 0.160 & 0.098 & 0.132 & 0.063 & 0.045 & 0.045 & 0.042 & \underline{0.036} & 0.038 & 0.041 & \textbf{0.030} \\ \cline{1-1} \cline{3-3} \hline 
70 & 50 & 0.063 & 0.154 & 0.082 & 0.119 & 0.051 & 0.043 & \textbf{0.036} & \textbf{0.036} & 0.039 & 0.040 & \underline{0.038} & 0.040 \\ \cline{1-1} \cline{3-3} \hline 
70 & 60 & 0.067 & 0.158 & 0.086 & 0.116 & \textbf{0.047} & 0.071 & \underline{0.050} & 0.055 & 0.056 & 0.051 & 0.063 & 0.054 \\ \cline{1-1} \cline{3-3} \hline 
80 & 50 & 0.063 & 0.150 & 0.090 & 0.111 & 0.040 & \textbf{0.035} & 0.044 & 0.039 & 0.038 & \underline{0.036} & \textbf{0.035} & \underline{0.036} \\ \cline{1-1} \cline{3-3} \hline 
80 & 60 & 0.054 & 0.160 & 0.081 & 0.104 & 0.039 & 0.038 & 0.035 & 0.031 & 0.034 & \underline{0.029} & 0.053 & \textbf{0.026} \\ \cline{1-1} \cline{3-3} \hline 
80 & 70 & 0.063 & 0.170 & 0.092 & 0.089 & \textbf{0.037} & 0.052 & \underline{0.045} & 0.047 & 0.054 & 0.067 & 0.049 & 0.050 \\ \cline{1-1} \cline{3-3} \hline 
90 & 50 & 0.056 & 0.146 & 0.090 & 0.116 & 0.046 & 0.038 & 0.031 & 0.041 & 0.041 & 0.035 & \underline{0.029} & \textbf{0.026} \\ \cline{1-1} \cline{3-3} \hline 
90 & 60 & 0.053 & 0.130 & 0.097 & 0.094 & 0.039 & 0.030 & 0.029 & 0.040 & 0.040 & \underline{0.028} & 0.029 & \textbf{0.025} \\ \cline{1-1} \cline{3-3} \hline 
90 & 70 & 0.047 & 0.140 & 0.095 & 0.087 & 0.034 & 0.042 & \underline{0.031} & \textbf{0.027} & \underline{0.031} & 0.041 & 0.035 & 0.037 \\ \cline{1-1} \cline{3-3} \hline 
100 & 60 & 0.051 & 0.143 & 0.085 & 0.096 & 0.039 & 0.031 & 0.032 & 0.028 & \textbf{0.025} & \underline{0.027} & 0.030 & 0.030 \\ \cline{1-1} \cline{3-3} \hline 
100 & 70 & 0.065 & 0.130 & 0.088 & 0.086 & 0.032 & \underline{0.029} & \textbf{0.027} & 0.032 & \textbf{0.027} & 0.031 & 0.032 & 0.035 \\ \cline{1-1} \cline{3-3} \hline 
110 & 60 & 0.053 & 0.127 & 0.099 & 0.103 & 0.038 & 0.028 & \underline{0.025} & 0.030 & 0.031 & 0.027 & \textbf{0.023} & 0.026 \\ \cline{1-1} \cline{3-3} \hline 
110 & 70 & 0.054 & 0.149 & 0.084 & 0.094 & 0.037 & \textbf{0.023} & \textbf{0.023} & 0.034 & 0.036 & 0.034 & 0.035 & \underline{0.031} \\ \cline{1-1} \cline{3-3} \hline 
120 & 70 & 0.056 & 0.126 & 0.084 & 0.097 & 0.033 & 0.036 & \textbf{0.027} & 0.036 & \underline{0.029} & 0.044 & 0.042 & 0.037 \\ \cline{1-1} \cline{3-3} \hline 
130 & 70 & 0.055 & 0.137 & 0.081 & 0.092 & \textbf{0.033} & 0.048 & 0.047 & 0.062 & \underline{0.036} & 0.045 & 0.052 & 0.038 \\ \cline{1-1} \cline{3-3} \hline 
 & AVG & 0.064 & 0.154 & 0.092 & 0.113 & 0.048 & 0.045 & \underline{0.041} & 0.044 & 0.043 & 0.042 & 0.044 & \textbf{0.039} \\ \hline 
\end{tabular}}%
\end{table*}

We show the EER results for various choices of $w$ and $w_\textrm{in}$ in Table~\ref{tab:eer}. The values in each cell represent the average EER across all test samples for all users for the corresponding choices of $w$, $w_\textrm{in}$, and experiment. For our method, i.e., \texttt{3Dfrom2D\_WESHKA}, we obtain EER values ranging from 0.025 to 0.060. The best performance is obtained where the value of EER is lowest at 0.025, for $w$ of 90 and $w_\textrm{in}$ of 60. 

Observing Table~\ref{tab:eer} reveals that, across various combinations of $w$ and $w_\textrm{in}$, the lowest Equal Error Rate (EER) values, denoted in bold font, tend to cluster within the middle and lower sections of the table. We observe the following outcomes:
\begin{enumerate}
    \item The proposed approach exhibits superior performance compared to the approach of Li et al~\cite{li2024using},
    \item Predicting 3D trajectories from 2D body joints and subsequently conducting identity authentication outperforms the methods of predicting 2D trajectories and then performing identity authentication, and
    \item The incorporation of additional 2D body joints improves the model's overall authentication performance.
\end{enumerate}
Among the 20 combinations of $w$ and $w_\textrm{in}$, our method outperforms the other approaches in 7 instances, indicated by achieving the lowest EER (in bold). Furthermore, our method ranks the second best in 4 out of the rest 13 combinations, as evidenced by the second-to-last lowest EER (values underscored). On average, across all 20 combinations and experiments, our approach achieves the best result, with an average EER value of 0.039, the lowest among all compared instances, as highlighted in bold in the last row in Table~\ref{tab:eer}.


Compared to the prior work of Li et al.~\cite{li2024using}, we observe that our approach outperforms their method for all choices of $w$ and $w_\textrm{in}$. 
It is worth noting that the approach of Li et al. uses 3D data at the input, and we incorporate part of the original 3D data in the authentication method to retain a signature of the user's original performance.  These aspects of using and encapsulating full 3D information enable their method to show a higher authentication success over other experiments, including \texttt{2Dfrom2D\_W} and \texttt{2Dfrom2D\_WES} that perform authentication solely based on 2D trajectory predictions as well as \texttt{3Dfrom2D\_W} that predicts the 3D trajectory. However, when multi-joint input is combined with 3D trajectory prediction, our method shows a performance boost, demonstrating the benefit of using the user's external video to gain an enhanced knowledge of their behavior while still leveraging the benefits of 3D device trajectories for VR authentication. The average EER drop between our approach and Li et al.~\cite{li2024using} is 0.025 amounting to a 39.1\% reduction on average, and the highest drop is acquired at $w$ of 70 and $w_\textrm{in}$ of 40 at 0.040 amounting to a 57.1\% reduction in EER.

\section{Conclusion and Future Work}
\label{sec:conclusion}

We present an approach to perform behavior-based biometric authentication of users in VR environments by using 2D data from external cameras to predict 3D trajectories that encapsulate current and future motion, and use the 3D trajectories in authentication. Our method incorporates the advantages of combining future motion with past motion demonstrated in Li et al.~\cite{li2024using}, while overcoming the deficiencies in Li et al. in modeling user behavior by incorporating information from multiple joints of the dominant arm and leg. Using 2D video enables us to acquire access to multiple joints by applying joint extraction algorithms such as OpenPose. Our method outperforms the work of Li et al.~\cite{li2024using} and baselines that use a single input and/or predict 2D output.

In this paper, we evaluate the advantages of using 6 joints to describe the motion of the right arm and leg. For the VR action considered in this work, particularly VR ball-throwing, the right arm and leg have the highest contribution toward the activity. Given that the arm and head motions are mapped to VR devices such as the headset and hand controllers, they are likely to have a high contribution in several VR activities such as pointing, object selection and movement, writing using a virtual pen, and advanced tasks such as gameplay, VR-based drone navigation, and VR driver training. As part of future work, we will explore joint tracks from both arms, legs, and the head to predict VR device headset and hand controller motions, and use them for authentication. Methods can investigate various viewpoints, e.g., front-facing viewpoints for tasks such as VR-based driver training. Additionally, incorporating more advanced 2D skeleton estimation techniques, such as RTMPose~\cite{jiang2023rtmpose}, will further refine motion analysis. Using higher-resolution cameras, methods can also investigate using 2D data on finger motion to generate 3D hand tracks for VR headsets that use controller-free hand tracking.


\bibliographystyle{IEEEtran}
\bibliography{template}

\begin{thebibliography}{10}
\providecommand{\url}[1]{#1}
\csname url@samestyle\endcsname
\providecommand{\newblock}{\relax}
\providecommand{\bibinfo}[2]{#2}
\providecommand{\BIBentrySTDinterwordspacing}{\spaceskip=0pt\relax}
\providecommand{\BIBentryALTinterwordstretchfactor}{4}
\providecommand{\BIBentryALTinterwordspacing}{\spaceskip=\fontdimen2\font plus
\BIBentryALTinterwordstretchfactor\fontdimen3\font minus \fontdimen4\font\relax}
\providecommand{\BIBforeignlanguage}[2]{{%
\expandafter\ifx\csname l@#1\endcsname\relax
\typeout{** WARNING: IEEEtran.bst: No hyphenation pattern has been}%
\typeout{** loaded for the language `#1'. Using the pattern for}%
\typeout{** the default language instead.}%
\else
\language=\csname l@#1\endcsname
\fi
#2}}
\providecommand{\BIBdecl}{\relax}
\BIBdecl

\bibitem{desselle2020augmented}
M.~R. Desselle, R.~A. Brown, A.~R. James, M.~J. Midwinter, S.~K. Powell, and M.~A. Woodruff, ``Augmented and virtual reality in surgery,'' \emph{Computing in Science \& Engineering}, vol.~22, no.~3, pp. 18--26, 2020.

\bibitem{munoz2022immersive}
J.~Mu{\~n}oz, S.~Mehrabi, Y.~Li, A.~Basharat, L.~E. Middleton, S.~Cao, M.~Barnett-Cowan, J.~Boger \emph{et~al.}, ``Immersive virtual reality exergames for persons living with dementia: User-centered design study as a multistakeholder team during the covid-19 pandemic,'' \emph{JMIR Serious Games}, vol.~10, no.~1, p. e29987, 2022.

\bibitem{mehrabi2022immersive}
S.~Mehrabi, J.~E. Mu{\~n}oz, A.~Basharat, J.~Boger, S.~Cao, M.~Barnett-Cowan, L.~E. Middleton \emph{et~al.}, ``Immersive virtual reality exergames to promote the well-being of community-dwelling older adults: Protocol for a mixed methods pilot study,'' \emph{JMIR Research Protocols}, vol.~11, no.~6, p. e32955, 2022.

\bibitem{karaosmanoglu2022canoe}
S.~Karaosmanoglu, L.~Kruse, S.~Rings, and F.~Steinicke, ``Canoe vr: An immersive exergame to support cognitive and physical exercises of older adults,'' in \emph{CHI Conference on Human Factors in Computing Systems Extended Abstracts}.\hskip 1em plus 0.5em minus 0.4em\relax New York, NY: ACM, 2022, pp. 1--7.

\bibitem{minnekanti2024classesinvr}
J.~Minnekanti, N.~K. Banerjee, and S.~Banerjee, ``Classesinvr: An immersive vr-based classroom,'' in \emph{2024 IEEE International Conference on Artificial Intelligence and eXtended and Virtual Reality (AIxVR)}.\hskip 1em plus 0.5em minus 0.4em\relax IEEE, 2024, pp. 310--314.

\bibitem{gao2021digital}
H.~Gao, E.~Bozkir, L.~Hasenbein, J.-U. Hahn, R.~G{\"o}llner, and E.~Kasneci, ``Digital transformations of classrooms in virtual reality,'' in \emph{Proceedings of the 2021 CHI Conference on Human Factors in Computing Systems}, 2021, pp. 1--10.

\bibitem{bozkir2021exploiting}
E.~Bozkir, P.~Stark, H.~Gao, L.~Hasenbein, J.-U. Hahn, E.~Kasneci, and R.~G{\"o}llner, ``Exploiting object-of-interest information to understand attention in vr classrooms,'' in \emph{2021 IEEE Virtual Reality and 3D User Interfaces (VR)}.\hskip 1em plus 0.5em minus 0.4em\relax IEEE, 2021, pp. 597--605.

\bibitem{hasenbein2022learning}
L.~Hasenbein, P.~Stark, U.~Trautwein, A.~C.~M. Queiroz, J.~Bailenson, J.-U. Hahn, and R.~G{\"o}llner, ``Learning with simulated virtual classmates: Effects of social-related configurations on students’ visual attention and learning experiences in an immersive virtual reality classroom,'' \emph{Computers in Human Behavior}, vol. 133, p. 107282, 2022.

\bibitem{gawlik2020factors}
M.~Gawlik-Kobyli{\'n}ska, P.~Maciejewski, J.~Lebied{\'z}, and A.~Wysoki{\'n}ska-Senkus, ``Factors affecting the effectiveness of military training in virtual reality environment,'' in \emph{Proceedings of the 2020 9th International Conference on Educational and Information Technology}, 2020, pp. 144--148.

\bibitem{de2020use}
C.~de~Armas, R.~Tori, and A.~V. Netto, ``Use of virtual reality simulators for training programs in the areas of security and defense: a systematic review,'' \emph{Multimedia Tools and Applications}, vol.~79, pp. 3495--3515, 2020.

\bibitem{ahir2020application}
K.~Ahir, K.~Govani, R.~Gajera, and M.~Shah, ``Application on virtual reality for enhanced education learning, military training and sports,'' \emph{Augmented Human Research}, vol.~5, pp. 1--9, 2020.

\bibitem{rettinger2022defuse}
M.~Rettinger and G.~Rigoll, ``Defuse the training of risky tasks: Collaborative training in xr,'' in \emph{2022 IEEE International Symposium on Mixed and Augmented Reality (ISMAR)}.\hskip 1em plus 0.5em minus 0.4em\relax IEEE, 2022, pp. 695--701.

\bibitem{campbell2019uses}
A.~G. Campbell, T.~Holz, J.~Cosgrove, M.~Harlick, and T.~O’Sullivan, ``Uses of virtual reality for communication in financial services: A case study on comparing different telepresence interfaces: Virtual reality compared to video conferencing,'' in \emph{Future of Information and Communication Conference}.\hskip 1em plus 0.5em minus 0.4em\relax Berlin, Germany: Springer, 2019, pp. 463--481.

\bibitem{weise2016virtual}
S.~Weise and A.~Mshar, ``Virtual reality and the banking experience,'' \emph{Journal of Digital Banking}, vol.~1, no.~2, pp. 146--152, 2016.

\bibitem{odeleye2023virtually}
B.~Odeleye, G.~Loukas, R.~Heartfield, G.~Sakellari, E.~Panaousis, and F.~Spyridonis, ``Virtually secure: A taxonomic assessment of cybersecurity challenges in virtual reality environments,'' \emph{Computers \& Security}, vol. 124, 2023.

\bibitem{alsulaiman2008three}
F.~A. Alsulaiman and A.~El~Saddik, ``Three-dimensional password for more secure authentication,'' \emph{IEEE Transactions on Instrumentation and measurement}, vol.~57, no.~9, pp. 1929--1938, 2008.

\bibitem{alsulaiman2006novel}
------, ``A novel 3d graphical password schema,'' in \emph{2006 IEEE Symposium on Virtual Environments, Human-Computer Interfaces and Measurement Systems}.\hskip 1em plus 0.5em minus 0.4em\relax Piscataway, NJ: IEEE, 2006, pp. 125--128.

\bibitem{gurary2017leveraging}
J.~Gurary, Y.~Zhu, and H.~Fu, ``Leveraging 3d benefits for authentication,'' \emph{International Journal of Communications, Network and System Sciences}, vol.~10, no.~08, p. 324, 2017.

\bibitem{george2020gazeroomlock}
C.~George, D.~Buschek, A.~Ngao, and M.~Khamis, ``Gazeroomlock: Using gaze and head-pose to improve the usability and observation resistance of 3d passwords in virtual reality,'' in \emph{Intl. Conf. on Augmented Reality, Virtual Reality, and Computer Graphics}, 2020, pp. 61--81.

\bibitem{yu2016exploration}
Z.~Yu, H.-N. Liang, C.~Fleming, and K.~L. Man, ``An exploration of usable authentication mechanisms for virtual reality systems,'' in \emph{IEEE Asia Pacific Conference on Circuits and Systems}, 2016, pp. 458--460.

\bibitem{george2017seamless}
C.~George, M.~Khamis, E.~von Zezschwitz, M.~Burger, H.~Schmidt, F.~Alt, and H.~Hussmann, ``Seamless and secure vr: Adapting and evaluating established authentication systems for virtual reality,'' in \emph{NDSS}.\hskip 1em plus 0.5em minus 0.4em\relax San Diego, CA: NDSS, 2017, pp. 1--1.

\bibitem{olade2020exploring}
I.~Olade, H.-N. Liang, C.~Fleming, and C.~Champion, ``Exploring the vulnerabilities and advantages of swipe or pattern authentication in virtual reality (vr),'' in \emph{Proceedings of the 2020 4th International Conference on Virtual and Augmented Reality Simulations}.\hskip 1em plus 0.5em minus 0.4em\relax New York, NY: ACM, 2020, pp. 45--52.

\bibitem{funk2019lookunlock}
M.~Funk, K.~Marky, I.~Mizutani, M.~Kritzler, S.~Mayer, and F.~Michahelles, ``Lookunlock: Using spatial-targets for user-authentication on hmds,'' in \emph{Ext. Abstracts of the 2019 CHI Conference on Human Factors in Computing Systems}.\hskip 1em plus 0.5em minus 0.4em\relax New York, NY: ACM, 2019, pp. 1--6.

\bibitem{george2019investigating}
C.~George, M.~Khamis, D.~Buschek, and H.~Hussmann, ``Investigating the third dimension for authentication in immersive virtual reality and in the real world,'' in \emph{IEEE Conference on Virtual Reality and 3D User Interfaces}, 2019, pp. 277--285.

\bibitem{mustafa2018unsure}
T.~Mustafa, R.~Matovu, A.~Serwadda, and N.~Muirhead, ``Unsure how to authenticate on your vr headset? come on, use your head!'' in \emph{Proceedings of the Fourth ACM International Workshop on Security and Privacy Analytics}.\hskip 1em plus 0.5em minus 0.4em\relax New York, NY: ACM, 2018, pp. 23--30.

\bibitem{mmm2019vr}
A.~Kupin, B.~Moeller, Y.~Jiang, N.~K. Banerjee, and S.~Banerjee, ``Task-driven biometric authentication of users in virtual reality (vr) environments,'' in \emph{MultiMedia Modeling: 25th International Conference, MMM 2019, Thessaloniki, Greece, January 8--11, 2019, Proceedings, Part I 25}.\hskip 1em plus 0.5em minus 0.4em\relax Berlin, Germany: Springer, 2019, pp. 55--67.

\bibitem{pfeuffer2019behavioural}
K.~Pfeuffer, M.~J. Geiger, S.~Prange, L.~Mecke, D.~Buschek, and F.~Alt, ``Behavioural biometrics in vr: Identifying people from body motion and relations in virtual reality,'' in \emph{CHI Conference on Human Factors in Computing Systems}, 2019, pp. 1--12.

\bibitem{ajit2019combining}
A.~Ajit, N.~K. Banerjee, and S.~Banerjee, ``Combining pairwise feature matches from device trajectories for biometric authentication in virtual reality environments,'' in \emph{2019 IEEE International Conference on Artificial Intelligence and Virtual Reality (AIVR)}.\hskip 1em plus 0.5em minus 0.4em\relax Piscataway, NJ: IEEE, 2019, pp. 9--97.

\bibitem{miller2019}
R.~Miller, A.~Ajit, N.~K. Banerjee, and S.~Banerjee, ``Realtime behavior-based continual authentication of users in virtual reality environments,'' in \emph{2019 IEEE International Conference on Artificial Intelligence and Virtual Reality (AIVR)}.\hskip 1em plus 0.5em minus 0.4em\relax Piscataway, NJ: IEEE, 2019, pp. 253--2531.

\bibitem{mathis2020knowledge}
F.~Mathis, H.~I. Fawaz, and M.~Khamis, ``Knowledge-driven biometric authentication in virtual reality,'' in \emph{Extended Abstracts of the 2020 CHI Conference on Human Factors in Computing Systems}, 2020, pp. 1--10.

\bibitem{mathis2020rubikauth}
F.~Mathis, J.~Williamson, K.~Vaniea, and M.~Khamis, ``Rubikauth: Fast and secure authentication in virtual reality,'' in \emph{Ext. Abstracts of the 2020 CHI Conference on Human Factors in Computing Systems}.\hskip 1em plus 0.5em minus 0.4em\relax New York, NY: ACM, 2020, pp. 1--9.

\bibitem{mathis2020fast}
F.~Mathis, J.~H. Williamson, K.~Vaniea, and M.~Khamis, ``Fast and secure authentication in virtual reality using coordinated 3d manipulation and pointing,'' \emph{ACM Transactions on Computer-Human Interaction}, vol.~6, no.~1, pp. 1--1, Jan 2021.

\bibitem{miller2020within}
R.~Miller, N.~K. Banerjee, and S.~Banerjee, ``Within-system and cross-system behavior-based biometric authentication in virtual reality,'' in \emph{IEEE Conference on Virtual Reality and 3D User Interfaces Abstracts and Workshops}.\hskip 1em plus 0.5em minus 0.4em\relax IEEE, 2020, pp. 311--316.

\bibitem{olade2020biomove}
I.~Olade, C.~Fleming, and H.-N. Liang, ``Biomove: Biometric user identification from human kinesiological movements for virtual reality systems,'' \emph{Sensors}, vol.~20, no.~10, p. 2944, 2020.

\bibitem{miller2020personal}
M.~R. Miller, F.~Herrera, H.~Jun, J.~A. Landay, and J.~N. Bailenson, ``Personal identifiability of user tracking data during observation of 360-degree vr video,'' \emph{Scientific Reports}, vol.~10, no.~1, pp. 1--10, 2020.

\bibitem{miller2021using}
R.~Miller, N.~K. Banerjee, and S.~Banerjee, ``Using siamese neural networks to perform cross-system behavioral authentication in virtual reality,'' in \emph{IEEE Virtual Reality and 3D User Interfaces}.\hskip 1em plus 0.5em minus 0.4em\relax IEEE, 2021, pp. 140--149.

\bibitem{liebers2021understanding}
J.~Liebers, M.~Abdelaziz, L.~Mecke, A.~Saad, J.~Auda, U.~Gruenefeld, F.~Alt, and S.~Schneegass, ``Understanding user identification in virtual reality through behavioral biometrics and the effect of body normalization,'' in \emph{CHI Conference on Human Factors in Computing Systems}, 2021, pp. 1--11.

\bibitem{rack2023alyx}
C.~Rack, T.~Fernando, M.~Yalcin, A.~Hotho, and M.~E. Latoschik, ``Who is alyx? a new behavioral biometric dataset for user identification in xr,'' \emph{Frontiers in Virtual Reality}, vol.~4, p. 1272234, 2023.

\bibitem{liebers2024identifying}
J.~Liebers, S.~Brockel, U.~Gruenefeld, and S.~Schneegass, ``Identifying users by their hand tracking data in augmented and virtual reality,'' \emph{International Journal of Human--Computer Interaction}, vol.~40, no.~2, pp. 409--424, 2024.

\bibitem{li2024using}
M.~Li, N.~K. Banerjee, and S.~Banerjee, ``Using motion forecasting for behavior-based virtual reality (vr) authentication,'' in \emph{IEEE International Conference on Artificial Intelligence \& extended and Virtual Reality}.\hskip 1em plus 0.5em minus 0.4em\relax IEEE, 2024.

\bibitem{liebers2024kinetic}
J.~Liebers, P.~Laskowski, F.~Rademaker, L.~Sabel, J.~Hoppen, U.~Gruenefeld, and S.~Schneegass, ``Kinetic signatures: A systematic investigation of movement-based user identification in virtual reality,'' 2024.

\bibitem{giaretta2022security}
A.~Giaretta, ``Security and privacy in virtual reality--a literature survey,'' \emph{arXiv preprint arXiv:2205.00208}, 2022.

\bibitem{stephenson2022sok}
S.~Stephenson, B.~Pal, S.~Fan, E.~Fernandes, Y.~Zhao, and R.~Chatterjee, ``Sok: Authentication in augmented and virtual reality,'' in \emph{IEEE Symposium on Security and Privacy}.\hskip 1em plus 0.5em minus 0.4em\relax IEEE, 2022.

\bibitem{cao2017realtime}
Z.~Cao, T.~Simon, S.-E. Wei, and Y.~Sheikh, ``Realtime multi-person 2d pose estimation using part affinity fields,'' in \emph{IEEE Conference on Computer Vvision and Pattern Recognition}, 2017, pp. 7291--7299.

\bibitem{vaswani2017attention}
A.~Vaswani, N.~Shazeer, N.~Parmar, J.~Uszkoreit, L.~Jones, A.~N. Gomez, {\L}.~Kaiser, and I.~Polosukhin, ``Attention is all you need,'' \emph{Advances in Neural Information Processing Systems}, vol.~30, 2017.

\bibitem{jain2007handbook}
A.~K. Jain, P.~Flynn, and A.~A. Ross, \emph{Handbook of biometrics}.\hskip 1em plus 0.5em minus 0.4em\relax Berlin, Germany: Springer Science \& Business Media, 2007.

\bibitem{miller2022combining}
R.~Miller, N.~K. Banerjee, and S.~Banerjee, ``Combining real-world constraints on user behavior with deep neural networks for virtual reality (vr) biometrics,'' in \emph{2022 IEEE Conference on Virtual Reality and 3D User Interfaces (VR)}.\hskip 1em plus 0.5em minus 0.4em\relax Piscataway, NJ: IEEE, 2022, pp. 409--418.

\bibitem{liebers2021using}
J.~Liebers, P.~Horn, C.~Burschik, U.~Gruenefeld, and S.~Schneegass, ``Using gaze behavior and head orientation for implicit identification in virtual reality,'' in \emph{ACM Symposium on Virtual Reality Software and Technology}, 2021, pp. 1--9.

\bibitem{li2024evaluating}
M.~Li, N.~Zafar, N.~K. Banerjee, and S.~Banerjee, ``Evaluating deep networks for detecting user familiarity with vr from hand interactions,'' in \emph{2024 IEEE International Conference on Artificial Intelligence and eXtended and Virtual Reality (AIxVR)}.\hskip 1em plus 0.5em minus 0.4em\relax IEEE, 2024, pp. 226--230.

\bibitem{suzuki2023pinchkey}
M.~Suzuki, R.~Iijima, K.~Nomoto, T.~Ohki, and T.~Mori, ``Pinchkey: A natural and user-friendly approach to vr user authentication,'' in \emph{European Symposium on Usable Security}, 2023.

\bibitem{liebers2023exploring}
J.~Liebers, C.~Burschik, U.~Gruenefeld, and S.~Schneegass, ``Exploring the stability of behavioral biometrics in virtual reality in a remote field study: Towards implicit and continuous user identification through body movements,'' in \emph{ACM Symposium on Virtual Reality Software and Technology}, 2023.

\bibitem{zhou2021informer}
H.~Zhou, S.~Zhang, J.~Peng, S.~Zhang, J.~Li, H.~Xiong, and W.~Zhang, ``Informer: Beyond efficient transformer for long sequence time-series forecasting,'' in \emph{Proceedings of the AAAI conference on artificial intelligence}, vol.~35.\hskip 1em plus 0.5em minus 0.4em\relax Washington, DC: AAAI, 2021, pp. 11\,106--11\,115.

\bibitem{wang2017time}
Z.~Wang, W.~Yan, and T.~Oates, ``Time series classification from scratch with deep neural networks: A strong baseline,'' in \emph{IEEE International Joint Conference on Neural Networks}, 2017, pp. 1578--1585.

\bibitem{jiang2023rtmpose}
T.~Jiang, P.~Lu, L.~Zhang, N.~Ma, R.~Han, C.~Lyu, Y.~Li, and K.~Chen, ``Rtmpose: Real-time multi-person pose estimation based on mmpose,'' \emph{arXiv preprint arXiv:2303.07399}, 2023.

\end{thebibliography}

\end{document}